%% file: main.tex
\documentclass[runningheads]{llncs}

\usepackage{eccv}

\usepackage{eccvabbrv}

\usepackage{graphicx}
\usepackage{booktabs}
\usepackage{multirow,multicol}
\usepackage[accsupp]{axessibility}  
\usepackage{wrapfig,floatflt,graphicx}

\usepackage[pagebackref,breaklinks,colorlinks]{hyperref}

\usepackage{orcidlink}
\begin{document}

\title{Social-MAE: Social Masked Autoencoder for Multi-person Motion Representation Learning} 

\titlerunning{Social-MAE}

\author{Mahsa Ehsanpour\inst{1} \and
Ian Reid\inst{1,3} \and
Hamid Rezatofighi\inst{2}}

\authorrunning{M.~Ehsanpour et al.}

\institute{The University of Adelaide \and Monash University \and
Mohammad Bin Zayed University of Artificial Intelligence
\\
\email{Mahsa.ehsanpour@adelaide.edu.au}}

\maketitle

\begin{abstract}
  For a complete comprehension of multi-person scenes, it is essential to go beyond basic tasks like detection and tracking. Higher-level tasks, such as understanding the interactions and social activities among individuals, are also crucial. Progress towards models that can fully understand scenes involving multiple people is hindered by a lack of sufficient annotated data for such high-level tasks. To address this challenge, we introduce Social-MAE, a simple yet effective transformer-based masked autoencoder framework for multi-person human motion data. The framework uses masked modeling to pre-train the encoder to reconstruct masked human joint trajectories, enabling it to learn generalizable and data efficient representations of motion in human crowded scenes. Social-MAE comprises a transformer as the MAE encoder and a lighter-weight transformer as the MAE decoder which operates on multi-person joints' trajectory in the frequency domain. After the reconstruction task, the MAE decoder is replaced with a task-specific decoder and the model is fine-tuned end-to-end for a variety of high-level social tasks. Our proposed model combined with our pre-training approach achieves the state-of-the-art results on various high-level social tasks, including multi-person pose forecasting, social grouping, and social action understanding. These improvements are demonstrated across four popular multi-person datasets encompassing both human 2D and 3D body pose.
  \keywords{Social Masked Auto-Encoder, Unsupervised Pre-Training, Multi-Person Motion Representation.}
\end{abstract}

\input{intro}
\input{related_work}
\input{method}
\input{experiment}
\input{conclusion}


%
%
\bibliographystyle{splncs04}
\bibliography{egbib}

\newpage
\input{supp}

\end{document}

%% file: intro.tex
\vspace{-.5em}
\section{Introduction.}
\vspace{-.5em}
To achieve a thorough understanding of multi-person scenes, it is essential to tackle high-level tasks that demand precise comprehension of fine-grained human motion and social human behavior. These tasks include forecasting human global trajectories~\cite{gupta2018social,kosaraju2019social}, predicting multi-person local motion trajectories~\cite{adeli2021tripod,wang2021multi,xustochastic}, estimating body language and emotions~\cite{luo2020arbee}, as well as detecting social interactions~\cite{ehsanpour2020joint,han2022panoramic,li2022self} and actions within human crowded environments~\cite{ehsanpour2022jrdb}. If performed well, these tasks could serve as a promising tool for applications such as self-driving cars, motion-based human-robot interaction, sports performance analysis, and healthcare monitoring.

In order to accurately model these downstream tasks, it is essential to consider human body pose and motion information in multi-person and social scenarios, where interactions between individuals in crowded scenes play a significant role. By taking into account this information, models can better capture the nuances of human behavior and interactions, leading to more accurate and effective results in a variety of applications~\cite{wang2021multi,xustochastic}.

While estimating 2D and 3D human body pose from images and videos can be reliably achieved due to recent methodological advancements~\cite{cao2017realtime,fang2017rmpe,pishchulin2016deepcut,wang2020deep,xu2022vitpose} and the availability of large-scale human pose datasets~\cite{guler2018densepose,mahmood2019amass,von2018recovering}, acquiring large-scale annotation for higher-level, pose-dependent, social tasks is often challenging and requires a significant amount of effort and resources. Due to the mentioned difficulty, such datasets are not easy to obtain and supervised training alone is not enough to achieve the desired level of performance and generalizability in the downstream tasks. To overcome this challenge, pre-training preferably in an unsupervised manner is a crucial phase for models to generalize well. 

Following the great success of masked modeling in NLP, vision and beyond~\cite{devlin2018bert,feichtenhofer2022masked,he2022masked,tong2022videomae}, we propose to utilize masked modeling on multi-person human 2D and 3D body pose and motion data as our unsupervised pre-training task. We propose Social-MAE, a simple yet effective transformer-based masked autoencoder framework that can be adapted to both 2D and 3D human motion data as input and a variety of downstream tasks. The architecture of Social-MAE includes a transformer serving as the MAE encoder and a lighter-weight transformer acting as the MAE decoder. These components are specifically designed to operate on the trajectory of multi-person joints as input tokens.

Given the full set of input tokens, we randomly mask a subset and train Social-MAE to reconstruct masked tokens~(human joint trajectories). The social-MAE encoder operates only on unmasked tokens and the decoder reconstructs the full set of tokens given the latent representation of unmasked ones from the encoder and an initialization regarding the masked tokens. Instead of predicting multi-human joints' trajectory~(full set of tokens) in Cartesian coordinates directly, we encode human motion into the frequency domain via utilizing Discrete Cosine Transform~(DCT) to take advantage of the periodic and continuous nature of motion data. Once trained, the social-MAE decoder is replaced with a task-specific decoder and the model can be fine-tuned end-to-end for a variety of pose-dependent, high-level tasks and datasets in a fully-supervised manner.

Unsupervised masked pre-training on multi-person human motion data provides the encoder with a richer understanding of the underlying structure and motion patterns of the human body, particularly in the context of interactions among individuals. This allows the encoder to learn more generalizable and data efficient representations of motion, which can be applied across different downstream tasks. Furthermore, by not relying on specific labeled data for supervised training, the pre-trained encoder is less likely to overfit to the training data, which can result in better generalization performance. Our proposed Social-MAE stands out as one of the few studies that explores the efficacy of masked autoencoder modeling when dealing with \emph{a set of spatio-temporal data}, such as multi-human body motion trajectories.
Social-MAE sets a new state-of-the-art in three different tasks namely multi-person future motion forecasting~\cite{wang2021multi,xustochastic}, social grouping~\cite{han2022panoramic,li2022self} and social action detection~\cite{ehsanpour2022jrdb}.

In summary, the main contribution of this paper is two-fold:
\begin{itemize}
    \item {We present Social-MAE, a simple yet effective human joint trajectory masked autoencoder that unleashes the potential of self-supervised multi-person joints' trajectory reconstruction for a variety of downstream tasks. Our Social-MAE aligns with NLP, Images, and Videos' findings on masked modeling. We have demonstrated that masking and reconstructing multi-human joints' trajectory is a straightforward pre-training approach that yields a satisfactory solution to pre-train models for a variety of social downstream tasks.
    }
    \newline
    \item {Our proposed model outperforms those that were trained from scratch on various downstream tasks, namely, multi-person pose and motion forecasting, social grouping, and activity understanding and sets a new state-of-the-art on all the mentioned pose-dependent, high-level tasks. These enhancements have been showcased on four widely-used datasets that cover both 2D and 3D body pose for mult-person scenarios.}
\end{itemize}
\vspace{-1.1em}

%% file: related_work.tex
\vspace{-.1em}
\section{Related Work.}
\vspace{-.5em}
\textbf{High-Level Pose-Dependent Social Tasks.}

\textit{Multi-person human pose and motion modeling.~}Trajectory forecasting in social context has been widely studied in the literature~\cite{gupta2018social,kosaraju2019social,kothari2021interpretable,mangalam2020not,sadeghian2019sophie,sun2020recursive,tsao2022social}. Social trajectory forecasting only predicts coarse-grained and global multi-person future motion. To predict fine-grained human body motion in social context, Adeli et al.~\cite{adeli2020socially,adeli2021tripod} performs multi-person pose forecasting using pose and visual data and proposes a Social Motion Forecasting~(SoMoF) Benchmark. Following~\cite{adeli2020socially}, other works have focused to address the task~\cite{guo2022multi,wang2021multi,xustochastic}. ~\cite{guo2022multi} performs two-person motion forecasting by utilizing cross attention.~\cite{wang2021multi,xustochastic} propose a multi-range transformer-based framework which performs multi-person pose forecasting in deterministic and stochastic manner, respectively; however, both are limited to predict the pose for just one person in the scene at a time and require multiple
passes to make prediction for a multi-person scene. Recently, Vendrow et al.~\cite{vendrow2022somoformer} proposed a transformer architecture for this task by utilizing joints' trajectory of all people in the scene as input tokens which is the closest to our proposed transformed-based architecture. It adopts a simple token masking strategy to introduce noise in input for better generalization in a fully-supervised manner. All of the listed studies employ customised and carefully designed architectures for this particular task and rely on fully annotated data (full-supervision) during the training process to perform well.

Our proposed Social-MAE architecture is versatile and can be readily adapted to a variety of downstream tasks including multi-person pose forecasting and is capable of handling both 2D and 3D input data. Additionally, it benefits from the proposed self-supervised pre-training strategy, enhancing its generalizability and performance after fine-tuning for the downstream tasks.

\textit{Social Grouping and Action Understanding.~}Social grouping involves the process of clustering individuals within a scene by analyzing and considering their interactions with one another. Prior works aim at detecting groups in the scene~\cite{choi2014discovering,ge2012vision} by utilizing small-scale datasets and by relying on hand-crafted rules applicable to specific type of groups such as conversational ones~\cite{hung2011detecting,patron2010high,setti2013multi,swofford2020improving}. In contrast to previous works which are task dependent and require domain expert knowledge to carefully design hand-crafted rules,~\cite{ehsanpour2020joint} extends the concept of grouping to more general types of interactions. Further,~\cite{ehsanpour2020joint} tackles social grouping and action detection in social context as complementary tasks. Following~\cite{ehsanpour2020joint},~\cite{ehsanpour2022jrdb,han2022panoramic} tackled social grouping and action detection simultaneously. All the mentioned deep methods utilize complex architectures to process video input data. Recently, Li et.~al.~\cite{li2022self} tackles only social grouping by utilizing pose data as input and employs a self-supervised strategy to train the model.~\cite{jahangard2023real} also utilizes trajectory information to learn social-groups.

On the contrary to most of the mentioned works, our proposed social-MAE architecture is simple and performs only on multi-person 2D and 3D pose data. In addition to multi-person pose forecasting, we adapted social-MAE to social grouping and action understanding tasks. Further, it vastly outperforms~\cite{li2022self} utilizing the same data modality in social grouping. 

\textbf{Masked Modeling.~}
The field of deep learning has experienced a tremendous growth in the scale and complexity of architectures used. Hence, pre-training plays a crucial role in mitigating the risk of overfitting. While historically, pre-training has commonly relied on transfer learning and by utlizing large-scale datasets~\cite{carreira2017quo,deng2009imagenet}, more recent trends have shown the value of self-supervised and semi-supervised pre-training approaches~\cite{assran2021semi,caron2021emerging,chen2020simple}. Most recently, masked modeling has emerged as a highly effective approach in NLP, vision and beyond. Initially, the masking was approached as a form of noise in denoised autoencoders that removed noise or filled in gaps with contextual information through the use of convolution~\cite{vincent2008extracting,vincent2010stacked}. Following the success of GPT~\cite{brown2020language,radford2019language} in NLP, iGPT\cite{chen2020generative} was proposed to predict over a sequence of pixels. More recently, transformer-based architectures~\cite{dosovitskiy2020image} has been widely used for masked modeling~\cite{bao2021beit,dong2021peco,he2022masked,wei2022masked,xie2022simmim,zhou2021ibot}. Following BERT~\cite{devlin2018bert} in NLP, BEiT~\cite{bao2021beit}, BEVT~\cite{wang2022bevt} and VIMPAC~\cite{tan2021vimpac} were proposed to learn image and video representation by discrete token prediction~\cite{ramesh2021zero}. Masked autoencoder~(MAE) frameworks have most recently been introduced for different types of data modalities such as image~\cite{he2022masked} and video~\cite{feichtenhofer2022masked,tong2022videomae}. The architecture of MAE is asymmetrical as it reconstructs the full input data using \emph{sparse observations}. The encoder is deep and efficient, as it only processes sparse input tokens, while the decoder is shallow. Consequently, the encoder learns rich representations of the data and proven to be effective as a pre-training approach for downstream image and video classification tasks. Sparse masked modeling has been investigated beyond vision including point cloud~\cite{liu2022masked,min2022voxel,pang2022masked,zhang2022point}, mesh~\cite{liang2022meshmae} and graph~\cite{chen2022graph,hou2022graphmae,li2022maskgae,tan2022mgae} data structures. Recently, Social-SSL~\cite{tsao2022social} has investigated dense masked modeling as a self-supervised pre-training approach for trajectory forecasting. Moreover, PoseBERT~\cite{baradel2022posebert} investigates dense masked modeling on single human pose and shape reconstruction. Neither of the works utilize the sparse MAE approach, instead opting for densely utilizing all the tokens, thereby missing out on sparse MAE advantages.~\cite{jiang2022dual} explores sparse MAE to reconstruct 3D body pose given partial 2D pose in multi-person scenarios on Shelf dataset~\cite{Chen_2020_CVPR}. MAE as a self-supervised pre-training approach for 2D and 3D multi-person pose data has not been investigated in this work.

Our proposed Social-MAE, investigates sparse masked modeling on 2D and 3D multi-human motion data which has not been widely studied in the literature. Further, it is employed as a promising self-supervised pre-training strategy for a variety of social and pose-dependent downstream tasks.

%% file: method.tex
\vspace{-.5em}
\section{Method.}
\vspace{-.5em}

In this section, we first present the proposed Social-MAE's autoencoder architecture and our self-supervised sparse masking strategy for processing and reconstructing 2D/3D multi-person body joint trajectories. We then proceed to describe the process of adapting and fine-tuning the pre-trained Social-MAE to each of the downstream tasks.

The following notation will be used throughout the remainder of this section. We assume an input sequence with $N$ people in motion, each composed of $J$ joints, in which the pose sequence of time duration $T$ for each person is denoted by $X_{1:T}^{n,j}=[x_1^{n,j},x_2^{n,j},~\text{...}, x_T^{n,j}]$ where $n\in\{1,~\text{...},N\}$, $j\in\{1,~\text{...},J\}$.
Variable length sequences and missing joints are handled by padding. Each $x_t^{n,j}$ therefore denotes the coordinates of the $j^{\mbox{th}}$ joint of the $n^{\mbox{th}}$ person at time $t$. Depending on the dataset these coordinates might be either 3D-Cartesian values, or 2D image coordinates.

\begin{figure*}[tbh]
   \centering
   \includegraphics[width=\textwidth]{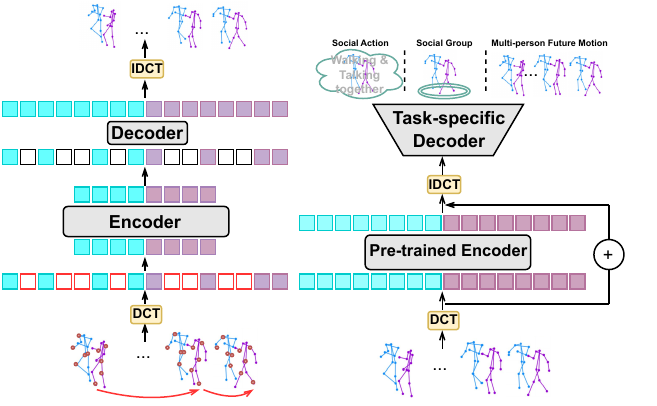}
   \caption{On the left, the initial tokens to the social-MAE are joints' trajectory of all people in the scene. We randomly mask a subset of input tokens. The encoder performs on a small subset of input tokens~(unmasked tokens only). The small decoder is  then applied on the latent representation of unmasked tokens as well as the masked ones to reconstruct the original input joints' trajectory. On the right and after pre-training, the decoder is replaced with a task-specific decoder for each downstream task and the pre-trained encoder is applied on the full-set of multi-person joints' trajectory.}
   \label{fig:framework}
    \vspace{-1em}
   \end{figure*}

\textbf{Social Masked Autoencoder.~}MAE is a transformer-based asymmetric autoencoder that reconstructs the full input signal given sparse and partially visible observations~\cite{feichtenhofer2022masked,he2022masked,tong2022videomae}. Similarly, our proposed Social-MAE is an asymmetric masked autoencoder that reconstructs the original multi-person joint trajectories given partial joints' trajectories that have not been masked.

\textbf{Input Formulation.~}The input tokens to the Social-MAE are joints' trajectories of multiple people in the scene indicated as $X_{1:T}^{n,j},\forall n \in N,~\forall j \in J$. Before masking, we make adjustments to the joint trajectory queries by removing global translation through subtracting the pelvis joint coordinate from the last-known frame for each individual. We encode human joints trajectory into the frequency domain using Discrete Cosine Transform~(DCT) to take advantage of the periodic and continuous nature of motion data. The global position for each person is then encoded via a learned embedding and combined with the joint trajectory representation. Personal identity and joint type~(pelvis, left-knee, etc.) are also encoded using learnable embeddings that are added to the queries.

\textbf{Masking.~}Considering all the joint trajectories of people in the scene as the initial tokens, we randomly mask a subset of them. In particular, the entire trajectory of a joint of a person~$X_{1:T}^{n,j}$ is masked, which we call Tube Masking. This masking approach prevents information leakage that occurs due to the temporal correlation present in the data. By randomly masking the complete trajectory of joints, the model is exposed to situations where certain joints' information is entirely missing. This helps prevent the model from relying solely on the temporal patterns and correlations between timesteps during training.

\textbf{Architecture.~}Social-MAE is an asymmetric masked autoencoder framework consisting of a transformer as the MAE encoder and a shallow transformer serving as the MAE decoder. After masking, the remaining  joints trajectory belonging to multiple people are fed to the MAE encoder. Using the latent representation obtained from the MAE encoder and the initialized tokens for the masked trajectories, the lightweight decoder reconstructs the complete original trajectory of multi-person joints as depicted in Fig.~\ref{fig:framework}-left. The reconstruction is performed using mean square error~(MSE) between the predicted and groundtruth joints trajectory as the training objective. The decoder is only utilized for the reconstruction task and will be replaced by a task-specific decoder in the fine-tuning phase for each downstream task as depicted in Fig.~\ref{fig:framework}-right. The MAE encoder pre-trained with the reconstruction task will be utilized to produce multi-person motion representations for each downstream task.

In the following, we will explain the process of adapting Social-MAE to each downstream task. It should be noted that for all the tasks, the input consists of the complete set of multi-person joints trajectory without any masking. Additionally, the same pre-trained MAE encoder is employed as the multi-person joints trajectory encoder on either 2D or 3D data across all the tasks. To fine-tune the model end-to-end for each specific task, we solely substitute the MAE decoder with a task-specific decoder and loss function.

\textbf{Task~1: Social Pose Forecasting.~}The goal of this task is to forecast the future motion of multi-person joints given their motion history. More specifically, given the joints' trajectory history of people~$X_{1:T}^{n,j}$ in the scene, we wish to predict their motion for~$\tau$ number of timesteps in the future~$X_{T+1:T+\tau}^{n,j}$. To this end, we utilize the pre-trained MAE encoder as our multi-person joints' trajectory encoder. We simply fine-tune the pre-trained encoder to predict the future motion for each input query by utilizing mean square error between the predicted and groundtruth future coordinates as indicated in Eq.~\ref{eq:L_PF} and without requiring a task-specific decoder. L denotes the number of encoder layers and we encourage the model to converge faster by utilizing an auxiliary loss at each encoder layer.\vspace{-.5em}
\begin{equation}
\vspace{-.3em}
\label{eq:L_PF}
    \mathcal{L}=\sum_{l=1}^{L} \sum_{n=1}^{N}\sum_{j=1}^{J}\lambda_{L} ||X_{T+1:T+\tau}^{n,j} - \hat X_{T+1:T+\tau}^{n,j}||^2
\end{equation}

\textbf{Task~2: Social Grouping.~}The objective of this task is to assign individuals to sub-groups by considering their interactions or contributions towards a common activity/a shared goal~\cite{ehsanpour2020joint,ehsanpour2022jrdb}. 
Given the joints' trajectory history of people~$X_{1:T}^{n,j}$ in the scene, we aim to predict the social group matrix $A_{N\times N} \in \{0,1\}$ where $A_{i,j}$ indicates whether $i$th and $j$th people belong to the same social group or not represented as 1 or 0, respectively, following previous methods~\cite{ehsanpour2020joint,ehsanpour2022jrdb}. 
Given the latent representation for each person from the pre-trained encoder, the pair-wise embedding distance is calculated between the people in the scene. The pair-wise embedding distance is concatenated with the pair-wise history trajectory distance. The obtained feature is passed through an MLP to predict the social group matrix, $A_{N\times N}$. Further, another MLP is utilized on top of the features to predict the number of social groups. We follow the training objective as in~\cite{ehsanpour2022jrdb} indicated in Eq.~\ref{eq:L_G}. A binary cross entropy loss is utilized between the elements of predicted~$A_{N\times N}$ and gruondtruth~$\hat A_{N\times N}$ denoted by $\mathcal{L}_{BCE}$. Since the number of connected components~(social groups) in the groundtruth matrix~$\hat A$ is equal to the number of zero eigenvalues of its laplacian matrix~$\mathrm{\hat L}$, we wish the laplacian matrix of~$A$ denoted by~$\mathrm{L}$ to have the same number of zero eigenvalues as in~$\mathrm{\hat L}$. To this end, $\mathcal{L}_{EIG}$ is utilized. To learn the number of social groups, $\mathcal{L}_{MSE}$ is utilized between the groundtruth and predicted number of social groups. 
\begin{equation}
\vspace{-.3em}
\label{eq:L_G}
    \mathcal{L}=\lambda_{1}\mathcal{L}_{BCE}(A,\hat A)+\lambda_{2}\mathcal{L}_{EIG}(\mathrm{L},\mathrm{\hat L})+\lambda_{3}\mathcal{L}_{MSE}(C, \hat C)
\end{equation}
Please refer to supp. material for more details regarding the social grouping loss function.

\textbf{Task~3: Social Action Understanding.~}This task involves action detection for each individual in human-crowded environments by modeling interactions between individuals. The output of the model is one pose-based action label~(walking, sitting, etc.) and an arbitrary number of interaction-based action labels~(talking to someone, listening to someone, etc.). Given the full set of input tokens to the pre-trained MAE encoder, we obtain the latent representation for each person in which social interactions are encoded. We utilize a MLP on top the representations to predict the set of individual's actions. Since each individual has one pose-based action label~$\hat P$ and an arbitrary number of interaction-based action labels~$\hat I$, we utilize a cross entropy loss and a binary cross entropy loss to supervise the model for each type of action respectively as indicated in Eq.~\ref{eq:L_act}. 
\begin{equation}
\vspace{-.8em}
\label{eq:L_act}
    \mathcal{L}=\lambda_{1}\mathcal{L}_{CE}(P,\hat P)+\lambda_{2}\mathcal{L}_{BCE}(\mathrm{I},\mathrm{\hat I})
\vspace{-.3em}
\end{equation}

%% file: experiment.tex
\section{Experimental Results.}
\vspace{-.7em}
  We demonstrate that our proposed social masked autoencoder for multi-person joints trajectory serves as a successful self-supervised pre-training strategy for high-level, pose-dependent tasks and sets a new state-of-the-art on all the downstream tasks across 4 datasets (indicated by \textbf{S-MAE} in Tabs.~[1-4]). Further, when trained from scratch in a supervised manner, we show that our proposed model~(baseline) either outperforms or is on par with the previous state-of-the-arts in multi-person pose forecasting, social grouping and action detection~(indicated as \textbf{Ours-B} in Tabs.~[1-4]).\vspace{-.8em} 
\subsection{Implementation Details.~}
\vspace{-.5em}
The MAE's transformer encoder consists of 6 layers, 8 attention heads, and a hidden dimension of 1024. The shallow MAE's transformer decoder consists of 3 layers, 4 attention heads and a hidden dimension of 1032. We use the masking ratio of 0.5 over the full set of tokens. We train the reconstruction model for 800 epochs with an initial learning rate of 0.0001 using ADAM optimizer and the learning rate decays to 0.001. We utilize learnable embeddings for joint type and person identity with the dimension of 1024 which are added to the tokens. Further, we concatenate a learnable embedding to the tokens which indicates the person's global position with the dimension of 8. We utilize the same architecture and hyper-parameters to pre-train the encoder for all the datasets and downstream tasks. For each downstream task the pre-trained encoder is kept the same and the transformer decoder is replaced with a task-specific decoder. For all the social tasks, the model is fine-tuned end-to-end for 256 epochs. All models are fine-tuned with an initial learning rate of 0.001 using ADAM optimizer and the learning rate decays to 0.01. For more details regarding the architecture of task-specific decoders, please refer to the supp. material.
\vspace{-.3em}
\subsection{Multi-person Pose Forecasting.}
\vspace{-.3em}
\textbf{Metrics.~}Following the prior works~\cite{mao2020history,wang2021multi}, we report Mean Per Joint Position Error~(MPJPE) and Visibility-Ignored Metric~(VIM). MPJPE measures the average Euclidean distances between groundtruth and predicted joint positions. First introduced in~\cite{adeli2021tripod} and used in the SoMoF benchmark, VIM calculates the mean 3D distance between the groundtruth and predicted joint positions after flattening together the joint and coordinate dimensions.

\textbf{Data.~}We utilize the same train and validation subsets as in~\cite{vendrow2022somoformer,wang2021multi}.

\textbf{3DPW~\cite{von2018recovering}.~}It provides real-world human motion sequences, containing over 60 video sequences. Since we used the SoMoF benchmark to evaluate our model, we use the SoMoF splits for 3DPW.

\textbf{AMASS~\cite{mahmood2019amass}.~}It provides a massive dataset of human motion capture sequences with over 40 hours of motion. During training, we use the CMU, BMLMovi and BMLRub subsets of this dataset. Since many of these sequences are single-person, we synthesize additional training data by mixing together sampled sequences to create multi-person training data.

\textbf{CMU-Mocap.~}It provides motion capture recording from 140 subjects performing various activities. We use training and testing sets derived by Wang et al.~\cite{wang2021multi} to train and evaluate our model.

\textbf{MuPoTS-3D~\cite{mehta2018single}.~}It provides 8,000 annotated frames of poses from 20 real-world scenes. We use this test set to evaluate our model performance.

In the following, we elaborate the multi-person pose forecasting quantitative results on CMU-Mocap, MuPoTS-3D and SoMoF test sets. It should be noted that since the multi-person pose forecasting datasets are in 3D space, our proposed architecture performs on 3D input tokens. Further, in the pre-training phase~(\textbf{S-MAE}), only the multi-person joints' history part is utilized for the reconstruction task which is of length 15 frames~(1 sec). For pre-training, we use an equivalent amount of data as other approaches use to supervise their models.

\textbf{SoMoF Benchmark Results.~}The SoMoF benchmark offers a standard for evaluating multi-person pose trajectory forecasting. Each sequence consists of 16 frames~(1.1 sec) of history as input, with the task of predicting the subsequent 14 frames~(0.9 sec). These frames contain the positions of joints for multiple individuals. The evaluation metric is the mean VIM across various future time steps. All models are trained using the 3DPW and AMASS datasets, which provide both multi-person and single-person data. Tab.~\ref{Tab:somof} presents a comparison of methods on the SoMoF test set, demonstrating that our model consistently outperforms all previous approaches.  

\textbf{CMU-Mocap and MuPoTS-3D Results.~}Using~\cite{wang2021multi} code and protocols to obtain train and test data, models in Tab.~\ref{Tab:PF} are trained using a synthesized dataset obtained from mixing motions in CMU-Mocap to obtain scences with 3 people, and are evaluated on CMU-Mocap and MuPoTS-3D. For multi-person pose forecasting, the input consists of 15 frames~(1 sec) of history, with the objective of predicting the subsequent 45 frames~(3 sec). The performance metric used is MPJPE, reported at 1, 2, and 3-second intervals in the future. For fairness in comparison, we train and evaluate each method using the same data. Our model consistently achieves better performance than other models on both datasets as indicated in Tab.~\ref{Tab:PF}.

\begin{table}[th]
\vspace{-1em}
\footnotesize
\caption{Experimental results in VIM on the SoMoF test set.}
  \label{Tab:somof}
  \centering
\begin{tabular}{lllllll}
\hline
           & \multicolumn{6}{c}{3DPW Prediction in Time}     \\ \cline{2-7} 
Method     & 100ms~$\downarrow$ & 240ms~$\downarrow$ & 500ms~$\downarrow$ & 640ms~$\downarrow$ & 900ms~$\downarrow$ & Overall~$\downarrow$ \\ \hline
SC-MPF~\cite{adeli2020socially} & 78.3   & 99.8  & 124.4  &  138.5 & 147.9  & 117.8 \\

TRiPOD~\cite{adeli2021tripod} & 30.3   & 51.8  & 85.1  &  104.8 & 146.3  & 83.6 \\
MRT~\cite{wang2021multi} & 22.9   & 42.3  & 79.4  & 99.0  & 137.9  & 76.3    \\
DuMMF~\cite{xustochastic} & 23.2   & 42.4  & 76.9  & 94.8  & 130.3  & 73.5    \\
FutureMotion~\cite{wang2021simple} & \textbf{9.5}   & 22.9  & 50.9  & 66.2  & 97.4  & 49.4    \\
\hline
\textbf{Ours-B} & 10.1  & 23.4  & 52.0  & 67.4  & 98.9  & 50.4        \\
\textbf{S-MAE}       & 9.6   & \textbf{22.4}  & \textbf{50.4}  & \textbf{64.8}  & \textbf{96.0}  &  \textbf{48.6}        \\
\hline
\end{tabular}
\end{table}
\begin{table}[th]
\footnotesize
\caption{MPJPE on CMU-Mocap and MuPoTS-3D Test Set~(0.1m).}
  \label{Tab:PF}
  \centering
\begin{tabular}{lllllll}
\hline
       & \multicolumn{3}{l}{CMU-Mocap Test Set}     & \multicolumn{3}{l}{MuPoTS-3D Test Set} \\ \cline{2-7} 
Method & 1 sec~$\downarrow$ & 2 sec~$\downarrow$ & \multicolumn{1}{l|}{3 sec~$\downarrow$} & 1 sec~$\downarrow$       & 2 sec~$\downarrow$       & 3 sec~$\downarrow$      \\ \hline
HRI~\cite{mao2020history}    & 0.503 & 0.932 & \multicolumn{1}{l|}{1.422} & 0.261       & 0.469       & 0.714      \\
LTD~\cite{mao2019learning}    & 0.480 & 0.869 & \multicolumn{1}{l|}{1.181} & 0.191       & 0.337       & 0.466      \\
MRT~\cite{wang2021multi}    & 0.456 & 0.84 & \multicolumn{1}{l|}{1.11} & 0.206       & 0.393       & 0.574      \\
\hline
\textbf{Ours-B}    & 0.42 & 0.80 & \multicolumn{1}{l|}{1.06} & 0.173       & 0.305       & 0.423      \\
\textbf{S-MAE}    & \textbf{0.39} & \textbf{0.78} & \multicolumn{1}{l|}{\textbf{1.04}} & \textbf{0.165}       & \textbf{0.285}       & \textbf{0.388}      \\
\hline
\end{tabular}
\vspace{-1em}
\end{table}
In the following, we elaborate the social grouping and action detection quantitative results on JRDB-Act~\cite{ehsanpour2022jrdb} validation and test subsets. It should be noted that on these tasks, our proposed architecture performs on 2D pose~(input tokens). Further, in the pre-training phase~(\textbf{S-MAE}), we randomly sample sequences of 16 frames from the JRDB-Act train set for the reconstruction task. Further, for pre-training, we use an equivalent amount of data as Ours-B utilizes to supervise the model from scratch. For reporting on validation set, we utilize the groundtruth pose annotation as input, however, to report on test set, we utilize the pose estimation~\cite{fu2023improved} and tracking~\cite{he2021know} methods trained and test on JRDB~\cite{martin2021jrdb} as input.\vspace{-.8em}

\subsection{Social Grouping.}
\vspace{-.5em}
\textbf{Metrics.~}Following the prior works~\cite{ehsanpour2020joint,ehsanpour2022jrdb}, we report AP for groups with $1, 2, 3, 4, 5^+$ members indicated by $G1, G2, G3, G4, G5^+$ respectively. mAP as the main social grouping metric~\cite{ehsanpour2022jrdb} is reported which is the average of the reported APs for groups with different number of members.

\textbf{Data.~}We utilize the JRDB-Act~\cite{ehsanpour2022jrdb} train, validation and test splits to train and evaluate our model. JRDB-Act comprises of 54 videos that are split into 20 train, 7 validation, and 27 test non-overlapping videos. Following the common practice, we evaluate our model's performance on the key-frames, which are sampled every one second, resulting in 404 validation, and 1802 test samples.

\textbf{JRDB-Act Benchmark Results.~}We evaluate our model on both validation and test subsets of JRDB-Act on key-frame level. As indicated in Tab.~\ref{Tab:SG}, our base model~(Ours-B) trained from scratch, vastly outperforms~\cite{li2022self} using similar input data~(2D skeleton joint). Further, our pre-trained model~(S-MAE) outperforms all the existing methods that utilize visual data~\cite{ehsanpour2020joint,ehsanpour2022jrdb,han2022panoramic} in terms of mAP. This shows that our model performs more balanced between groups with different number of members compared to the existing works.

\begin{table}[th]
\vspace{-1em}
\footnotesize
\caption{Social grouping performance on JRDB-Act validation and test sets.}
\label{Tab:SG}
\centering
\begin{tabular}{lllllll|l}
\hline
Set                      & Method       & G1~AP~$\uparrow$ & G2~AP~$\uparrow$ & G3~AP~$\uparrow$ & G4~AP~$\uparrow$ & G5$^+$~AP~$\uparrow$  & mAP~$\uparrow$  \\ \hline \hline
\multirow{5}{*}{\rotatebox[origin=c]{90}{Validation}} & SS-Rel~\cite{li2022self}  & 3.1   & 25.0  & 17.5  & 45.6  & 25.2   & 23.3   \\
                            & Multi-Social~\cite{ehsanpour2020joint}  & 8.0   & 29.3  & 37.5  & 65.4  & 67.0      & 41.4 \\
                            & P-HAR~\cite{han2022panoramic}     & 52.0  & 59.2  & 46.7  & 46.6  & 31.1    & 47.1      \\
                            & JRDB-Act~\cite{ehsanpour2022jrdb}   & 81.4  & 64.8  & 49.1  & 63.2  & 37.2     & 59.2  \\
                            \cline{2-8}
                            & \textbf{Ours-B}  & 58.5  & 61.2  & \textbf{53.6}  & 85.5 & \textbf{71.7} &66.1   \\
                            & \textbf{S-MAE}   &\textbf{65.6}   &\textbf{64.8}   &53.3   &\textbf{86.0}  &70.1 &\textbf{68.0}  \\
                            \hline \hline
\multirow{4}{*}{\rotatebox[origin=c]{90}{Test}}       & Multi-Social~\cite{ehsanpour2020joint}  & 11.2  & 24.6  & 21.8  & 41.2  & 23.5     & 24.5  \\
                            & JRDB-Act~\cite{ehsanpour2022jrdb}  & 56.3  & 39.4  & 24.3  & 22.9  & 15.3      & 31.6 \\
                            \cline{2-8}
                            & \textbf{Ours-B} &21.1       &39.8       &\textbf{29.0}       &\textbf{44.7}       &\textbf{24.5}    &31.8          \\ 
                            & \textbf{S-MAE}  & \textbf{24.4}      & \textbf{41.3}      & 28.7      & 43.8      & 24.0   & \textbf{32.4}          \\ 
                            
                            \hline
\end{tabular}
\vspace{-1.5em}
\end{table}

\vspace{-1em}
\subsection{Social Action Detection.}
\vspace{-.5em}

\textbf{Metric.~}mAP is reported on the key-frames as the standard practice~\cite{ehsanpour2022jrdb,gu2018ava}.

\textbf{Data.~}We utilize the JRDB-Act~\cite{ehsanpour2022jrdb} subsets to train and evaluate our model. JRDB-Act contains over 2.8M action labels, making it one of the large-scale spatio-temporal social action datasets publicly available. The sequences in JRDB-Act are captured by a mobile robot and naturally include diverse levels of human population density and complex interactions. The average number of people per frame in JRDB-Act is 30, which is significantly higher than most popular action datasets making it a suitable dataset to study action understanding in the social context.\vspace{-.5em}

\begin{wraptable}{l}{0.45\textwidth}
\vspace{-1em}
  \centering
  \footnotesize
  \caption{Social Action Detection on JRDB-Act validation and test sets.}
  \label{Tab:ACT}
  \begin{tabular}{lll}
    \hline
    Set & Method & mAP~$\uparrow$\\
    \hline \hline
    \multirow{4}{*}{\rotatebox[origin=c]{90}{Validation}}
    & Multi-Social~\cite{ehsanpour2020joint} & 8.1 \\
    & JRDB-Act~\cite{ehsanpour2022jrdb} & 9.0 \\
    \cline{2-3}
    & \textbf{Ours-B} & \textbf{13.5} \\
    & \textbf{S-MAE} & \textbf{14.1} \\
    \hline \hline
    \multirow{4}{*}{\rotatebox[origin=c]{90}{Test}}
    & Multi-Social~\cite{ehsanpour2020joint} & 4.9 \\
    & JRDB-Act~\cite{ehsanpour2022jrdb} & 5.4 \\
    \cline{2-3}
    & \textbf{Ours-B} & \textbf{10.9} \\
    & \textbf{S-MAE} & \textbf{11.2} \\
    \hline
  \end{tabular}
\end{wraptable}

\noindent
\textbf{JRDB-Act Benchmark Results.~}We evaluate our model on both validation and test subsets of JRDB-Act on key-frame level. As indicated in Tab.~\ref{Tab:ACT}, by only utilizing multi-person motion data as input~(Ours-B), our proposed model greatly outperforms the existing models that utilize only visual data~\cite{ehsanpour2020joint} or a combination of visual data and bounding box proximity features~\cite{ehsanpour2022jrdb} for social action detection task. Further, our self-supervised pre-training approach~(multi-person motion reconstruction) shows to be effective in better understanding of actions in social context once fine-tuned with full supervision for the task.
\subsection{Ablation Studies.}
\vspace{-.5em}
In this section, we present ablation studies on different properties of our proposed Social-MAE including data efficiency, pre-training masking ratio and architecture design. We report the ablation study results on the SoMoF validation set in terms of VIM.

\begin{figure*}
\vspace{-1em}
   \centering
   \includegraphics[width=\textwidth]{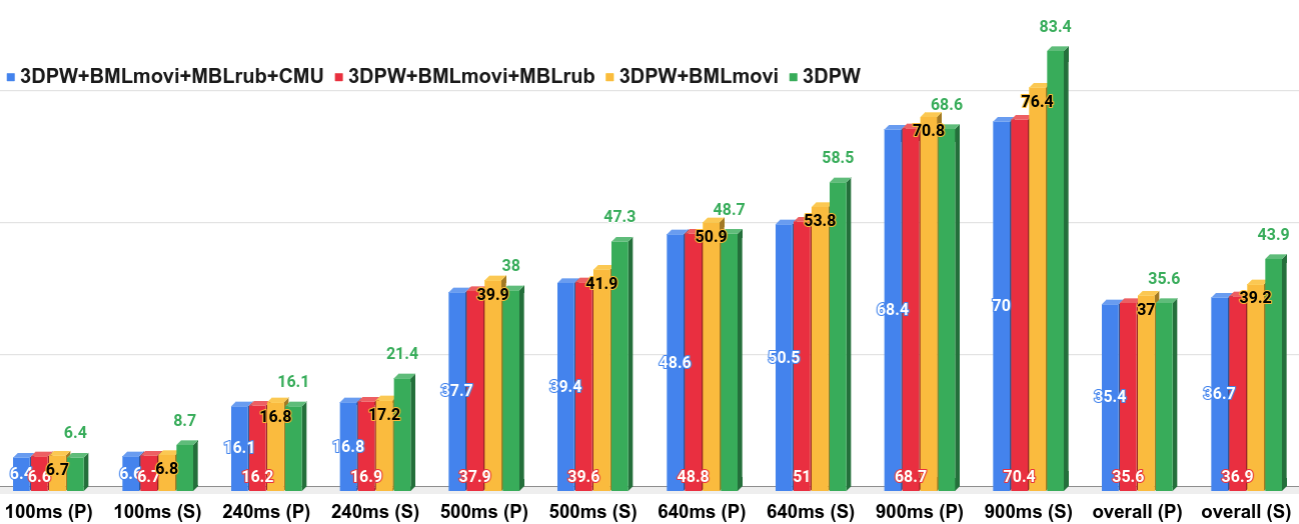}
   \caption{Ablation analysis to evaluate the performance error~(VIM) in relation to the number of available annotated data splits. We examined two scenarios: 1)~training our baseline model from scratch~(S). 2)~Fine-tuning our S-MAE model~(P). Note that S-MAE is pre-trained on all the data splits in the self-supervised manner. VIM is reported at each timestep on SoMoF validation set. Overall is the average of reported VIMs at different timesteps.}
   \label{fig:ab_1}
   \vspace{-1em}
   \end{figure*}

\textbf{Data Efficiency.~}In this part we conduct experiments on the effect of the amount of data utilized for the pre-training as well as fine-tuning phases.
To show the effect of the amount of data utilized for fine-tuning, we perform an ablation study on the performance error~(VIM) versus the number of the annotated data splits available for (a)~training our baseline~(from scratch), indicated by~(S) and (b)~fine-tuning our S-MAE, indicated by~(P) in Fig.~\ref{fig:ab_1}. Note that our S-MAE uses all the datasets~(without their annotated labels for the training task) in the self-supervised pre-training stage. Further, both models are trained/fine-tuned with the same amount of annotated data. For both settings, VIM on SoMoF validation set at different timesteps is reported in Fig.~\ref{fig:ab_1}. As observed, the pre-trained model demonstrates greater resilience to variations in the amount of fine-tuning data, particularly as the time step increases, compared to the model that does not utilize pre-training. Surprisingly, even with the utilization of only 3DPW data for fine-tuning the pre-trained model, the performance of the fine-tuned model is impressive when evaluated in terms of VIM on the SoMoF validation set. We believe this observation is due to the similarity of 3DPW data with SoMoF data.

\vspace{-.1em}
In the following, we further extend our ablation studies on the effect of the amount of data utilized for the pre-training task. In this study we pre-train the model with different amount of data and fine-tune the model with the full set of data available for multi-person pose forecasting task. As shown in Tab.~\ref{Tab:pre-data}, as the size of pre-training data decreases from top to bottom, the error increases and we observe performance drop.

\begin{table*}[t]
\vspace{-.5em}
\footnotesize
\caption{\textbf{Influence of the amount of pre-training data.}}
  \label{Tab:pre-data}
  \centering
\begin{tabular}{lllllll}
\hline
            & \multicolumn{6}{c}{SoMoF Prediction in Time~(ms)}    \\ \cline{2-7} 
\ Data & 100$\downarrow$ & 240$\downarrow$ & 500$\downarrow$ & 640$\downarrow$ & 900$\downarrow$ & Overall$\downarrow$ \\ \hline
\textbf{3DPW+BMLmovi+BMLrub+CMU}           & \textbf{6.4}   & \textbf{16.1}  & \textbf{37.7}  & \textbf{48.6}  & \textbf{68.4}  & \textbf{35.4}    \\
3DPW+BMLmovi+BMLrub          & 6.6   & 16.1  & 38.0  & 48.6  & 69.5  & 35.8    \\ 
3DPW+BMLmovi          & 6.7   & 16.9  & 39.9  & 51.1  & 70.8  &37.0     \\  \hline

\end{tabular}
\end{table*}

\textbf{Masking Ratio.~}As indicated in Tab.~\ref{Tab:mask_ratio}, masking ratio of 50\% gives the best performance on SoMoF validation set. Decreasing the mask ratio simplifies the reconstruction problem for the model which results in a a less generalizability and worse performance on the down-stream task. Increasing the masking ratio drops the performance as well. This finding supports the theory that the masking ratio is related to the redundancy of information within the data. Since multi-human joints trajectory data is sparse and contains less redundant information compared to images and videos, the optimal masking ratio is less~(50\%), compared to 90\% in videos~\cite{feichtenhofer2022masked,tong2022videomae} and 75\% in images~\cite{he2022masked}.  

\begin{table}[t]
\vspace{-.5em}
\footnotesize
\caption{\textbf{Influence of masking ratio.}}
  \label{Tab:mask_ratio}
  \centering

\begin{tabular}{lllllll}
\hline
            & \multicolumn{6}{c}{SoMoF Prediction in Time}    \\ \cline{2-7} 
Mask Ratio & 100ms~$\downarrow$ & 240ms~$\downarrow$ & 500ms~$\downarrow$ & 640ms~$\downarrow$ & 900ms~$\downarrow$ & Overall~$\downarrow$ \\ \hline
45\%          & 6.4   & 16.1  & 37.8  & 49.0  & 70.0  & 35.9    \\
\textbf{50\% }           & \textbf{6.4}   & \textbf{16.1}  & \textbf{37.7}  & \textbf{48.6}  & \textbf{68.4}  & \textbf{35.4}    \\
55\%            & 6.6   & 16.2  & 38.3  & 48.9  & 71.1.9  & 36.2    \\ 
60\%            & 6.7   & 16.5  & 38.9  & 49.6  & 71.3  & 36.6    \\ \hline
\end{tabular}
\vspace{-1em}
\end{table}

\textbf{Social-MAE Architecture.~}Our Social-MAE transformer encoder is inspired by~\cite{vendrow2022somoformer} with 6 layers, 8 attention heads and a hidden dimension of 1024. We follow the same strategy to add learnable embeddings to the input tokens as in~\cite{vendrow2022somoformer}. The encoder design choices have been ablated in~\cite{vendrow2022somoformer} and we observed similar results. Thus, the mentioned settings regarding the encoder are kept fixed in our experiments. We performed a study on the depth of the decoder~(the number of transformer layers) during the reconstruction phase. As indicated in Tab.~\ref{Tab:dec_layer}, a shallow decoder with 3 layers, gives the best performance on SoMoF validation set. This observation aligns with the image~\cite{he2022masked} and video~\cite{feichtenhofer2022masked,tong2022videomae} MAE counterparts in which the decoder is shallower than the encoder.

\begin{table}[th]
\vspace{-.8em}
\footnotesize
\caption{\textbf{Influence of decoder depth.}}
  \label{Tab:dec_layer}
  \centering
\begin{tabular}{lllllll}
\hline
            & \multicolumn{6}{c}{SoMoF Prediction in Time}    \\ \cline{2-7} 
\#DEC Layer & 100ms~$\downarrow$ & 240ms~$\downarrow$ & 500ms~$\downarrow$ & 640ms~$\downarrow$ & 900ms~$\downarrow$ & Overall~$\downarrow$ \\ \hline
2           & 6.5   & 16.2  & 37.9  & 48.7  & 69.1  & 35.7    \\
\textbf{3}           & \textbf{6.4}   & \textbf{16.1}  & \textbf{37.7}  & \textbf{48.6}  & \textbf{68.4}  & \textbf{35.4}    \\
4           & 6.6   & 16.2  & 37.9  & 48.6  & 68.9  & 35.6    \\ \hline
\end{tabular}
\vspace{-1em}
\end{table}
\vspace{-.8em}
\subsection{Qualitative Reslts.}
\begin{figure}[t]
\vspace{-.5em}
   \centering
   \includegraphics[width=\textwidth]{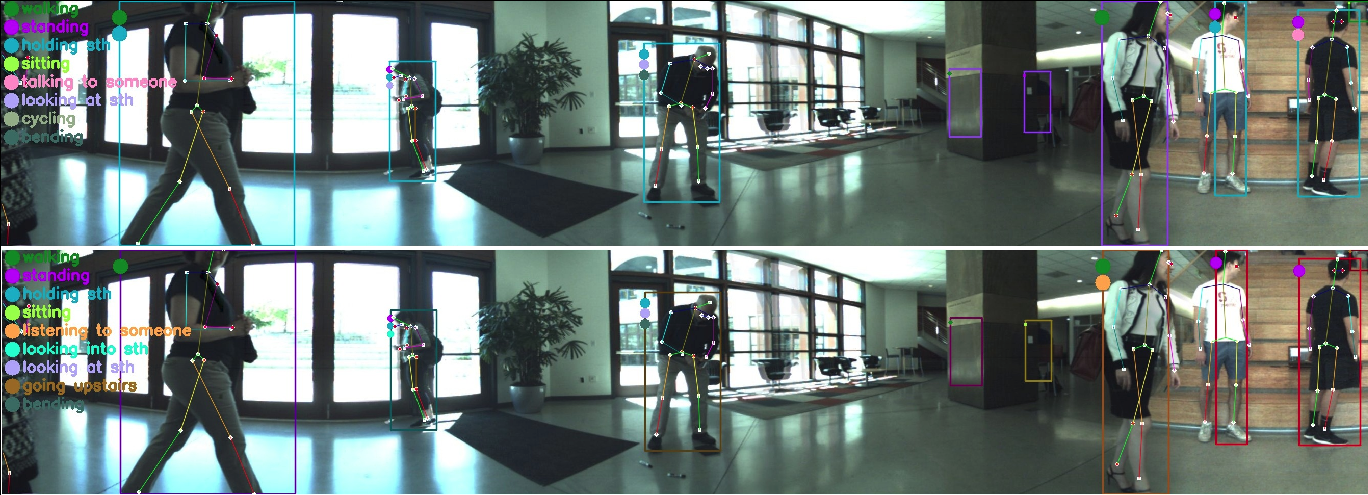}
   \caption{Social grouping and action understanding on JRDB-Act validation set. The top sub-figure indicates the S-MAE predictions and the bottom sub-figure shows the ground-truth. Social groups are indicated by bounding box colors and action labels are indicated by circles on left side of each box. Note that the same social group in the prediction and ground-truth sub-figures could have different colors.}\vspace{-2em}
   \label{fig:Q_res}
\end{figure}

\vspace{-.5em}
We have visualized the predictions of our proposed S-MAE model against the ground-truth on JRDB-Act validation set~\cite{ehsanpour2022jrdb} for the social grouping and social action understanding tasks in Fig.~\ref{fig:Q_res}. As shown, our proposed S-MAE qualitatively performs well on the social grouping and social action understanding tasks and matches the ground-truth in many cases. There are some failure cases of S-MAE in predicting the social actions. For instance, the model predicts false positive labels \textit{talking to someone, listening to someone} when multiple people are moving similarly and in each other's proximity. We believe that incorporating visual features would be a potential future direction to overcome this challenge.

\noindent
For more qualitative results please refer to the supp. material.

%% file: conclusion.tex
\vspace{-.5em}
\section{Impact.~}
\vspace{-.5em}
Our proposed approach is useful in many applications such as self-driving cars, human-robot interaction, sport analysis, and healthcare monitoring. Since our approach relies on utilizing real behavioral information and social interactions between humans as input, our concern is the exposure of sensitive identity details through body pose and motion. However, this risk is minimal as we do not utilize image and video modalities. A potential future direction is to leverage visual features in addition to motion information to enhance the comprehension of social actions. We hope this study encourages further research on pre-training approaches for pose-dependent, downstream social tasks.
\vspace{-.5em}
\section{Conclusion.~}
\vspace{-.5em}
The proposed Social-MAE presents an asymmetric transformer-based autoencoder framework for masked-modeling multi-person motion data. By employing masked modeling, the encoder is pre-trained to reconstruct masked human joint trajectories, enabling it to acquire more generalized motion representations. Subsequently, the MAE decoder is replaced with task-specific decoder and the model is fine-tuned end-to-end for each high-level, pose-dependent task. Through pre-training, the model exhibits superior performance in various high-level downstream tasks such as multi-person pose forecasting, social grouping, and action understanding when compared to supervised models trained from scratch across four datasets.

%% file: supp.tex
\input{imp_det}

\input{q_results}

\clearpage  

%
%

%% file: imp_det.tex
\noindent
\textbf{1~\quad Implementation Details.}
\noindent
As stated in the paper, for each social down-stream tasks, a task-specific decoder is utilized to be trained together with the pre-trained encoder in a fully-supervised end-to-end manner. We dive into more details of the task-specific decoders' architecture. Further, it should be noted that the pre-training architecture, strategy and hyper-parameters remains the same for all the down-stream tasks.

\textbf{Multi-person Pose Forecasting.~}Since the output of the encoder aligns with the expected down-stream task output~(joints trajectory of all the people in the scene), we simply fine-tune the pre-trained encoder with the MSE objective to accurately predict the future of multi-human joints' trajectory and without utilizing a task-specific decoder.

For the following down-strem tasks, in order to ensure a fair comparison, we employ identical training objectives as outlined in~\cite{ehsanpour2022jrdb} and adopt similar task-specific decoder architecture.

\textbf{Social Grouping.~}The decoder utilized to predict the social grouping matrix $A_{N\times N}$ is a MLP consisting of a stack of Linear layers and ReLU activation functions of size 16, 32, 128, 64, 8, 1, respectively. To predict the number of groups, we utilize a MLP on top of the embedding distances consisting of a stack of Linear layers and ReLU activation functions of size 8, 16, respectively. Further, another MLP is uitlized on top of the GIoU distances consisting of a stack of Linear layers and ReLU activation functions of size 8, 16, respectively. Lastly, the obtained feature maps are concatenated and a MLP is applied to predict the number of social group consisting of a stack of Linear layers and ReLU activation functions of size 16, 8, 1, respectively.

\textbf{Social Grouping Loss Function.}
As stated in \textbf{Task~2-Social Grouping} of Sec.~[3] in the paper, the number of connected components~(social groups) in the groundtruth matrix $\hat A$ is equal to the number of zero eigenvalues of its laplacian matrix $\hat L$. Thus, we want the laplacian matrix of $A$ denoted by $L$ to have the same number of zero eigenvalues as in $\hat L$. If $e$ is an eigenvector of $L^{T}L$~(in order to ensure that the matrix is symmetric) with the eigenvalue $\lambda$, it satisfies $L^{T}Le=\lambda e$. Since $e^{T}e=1$~(eigenvectors have unit-norm), multiplying both sides of the equation by $e^{T}$ yields $e^{T}L^{T}Le=\lambda$ and we want to consider eigenvalues of zero~($\lambda=0$). Given the groundtruth eigenvector $e$ corresponding to the zero eigenvalue, we define the loss as
\vspace{-.5em}
\begin{equation}
\mathcal{L}=e^{T}L^{T}Le; e^{T}L^{T}Le \geq 0
\end{equation}
To avoid the trivial solution $L=0$, a second term is added to maximize the projection of data along the directions orthogonal to $e$.
\vspace{-.5em}
\begin{equation}
\mathcal{L}=e^{T}L^{T}Le - \alpha tr(\bar{L}^{T}\bar{L})
\end{equation}
\vspace{-.5em}
and finally for numerical stability the second term is bounded in the range $[0,1]$ as \begin{equation}
\mathcal{L}_{EIG}=e^{T}L^{T}Le+\alpha\exp(-\beta tr(\bar{L}^{T}\bar{L}))
\end{equation}

Both $\alpha$ and $\beta$ are set to 1 in our experiments. 

\textbf{Social Action Understanding.~}To predict the individual set of actions for each person in the scene, we utilized a MLP that outputs the pose-based action labels consisting of a stack of Linear layers and ReLU activation functions of size 1024, 256, 64, 10, respectively. The final activation function is Softmax to predict a single pose-based action label for each individual during inference. Further, a MLP is utilized to output the set of interaction-based action labels consisting of a stack of Linear layers and ReLU activation functions of size 1024, 256, 64, 14, respectively. The final activation function is Sigmoid to predict an arbitrary number of interaction-based action labels for each individual during inference. We use the threshold of 0.6 to accept the predicted interaction-based actions. It should be noted that for all the self-supervised pre-training tasks the initial learning rate is 0.0001 which is decayed by 0.1 by the end of training. Similarly, for all the supervised fine-tuning tasks the initial learning rate is 0.001 which is decayed by 0.1 by the end of training.

%% file: q_results.tex
\noindent
\textbf{2~\quad Qualitative Results.}
\noindent
The predictions of S-MAE against the ground-truth on JRDB-Act validation set~\cite{ehsanpour2022jrdb} for the social grouping and social action understanding tasks is visualized in Fig.~[1-3] to showcase the correctness of the predictions as well as some failure cases of S-MAE. The input to the S-MAE model for both tasks, is the multi-person joints' trajectory for 16 frames~(the key-frame and 15 frames prior to that). In each figure, the top sub-figure is the output of S-MAE on key-frame and the bottom sub-figure is the key-frame's ground-truth. Social grouping is demonstrated by the bounding box color where boxes with the same color in each sub-figure belong to the same social group. Please note that it is not necessary for the same group in the prediction sub-figure and the ground-truth sub-figure to be depicted with the same color. Social action labels are shown with colored circles on each bounding box.

\begin{figure}
   \centering
   \includegraphics[width=\textwidth]{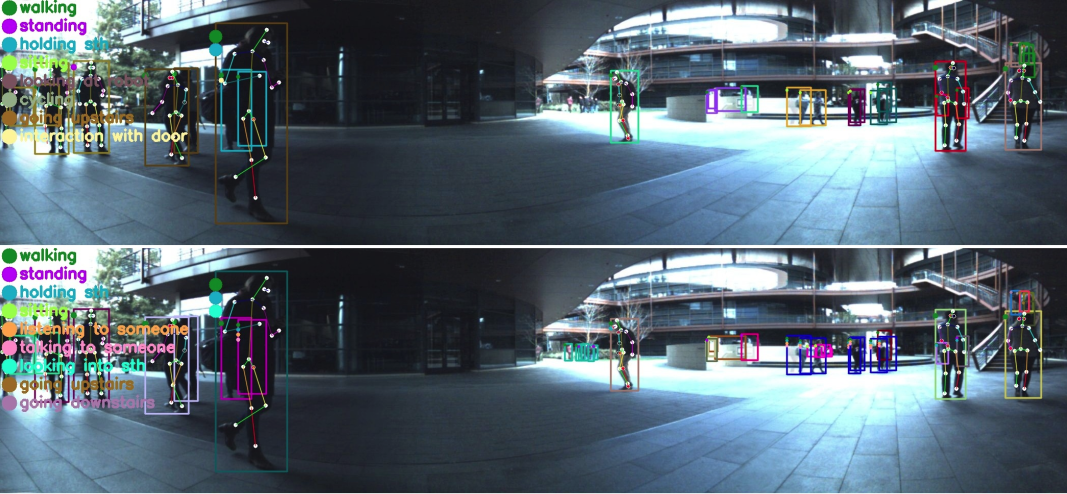}
   \vspace{-.5em}
   \hfill
   
   \includegraphics[width=\textwidth]{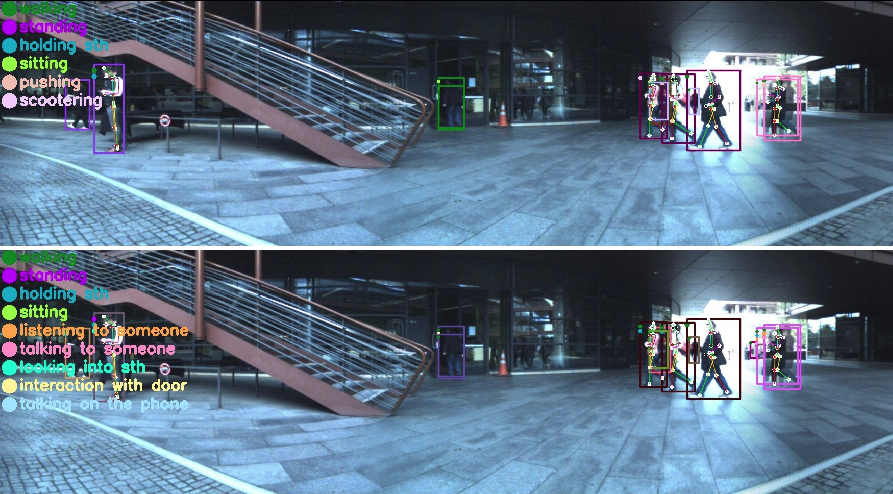}
   \vspace{-.5em}
   \hfill
   
   \includegraphics[width=\textwidth]{figs/vis_5.png}
   \vspace{-.5em}
   \caption{3 different samples of social grouping and action understanding on JRDB-Act validation set. The top sub-figure indicates the S-MAE predictions and the bottom sub-figure shows the ground-truth in each sample. Social groups are indicated by bounding box colors and action labels are indicated by circles on left side of each box. Note that the same social group in the prediction and ground-truth sub-figures could have different colors.}
   \label{fig:v1}
\end{figure}

As shown in Fig.~\ref{fig:v1}, S-MAE qualitatively performs well on the social grouping and social action understanding tasks and matches the ground-truth in many cases. There are some failure cases of S-MAE in predicting the social actions. For instance, the model predicts false positive labels \textit{talking to someone, listening to someone} when multiple people are moving similarly and in each other's proximity. We believe that incorporating visual features would be a potential future direction to overcome this challenge.

\begin{figure*}[h]
   \centering
   \includegraphics[width=\textwidth]{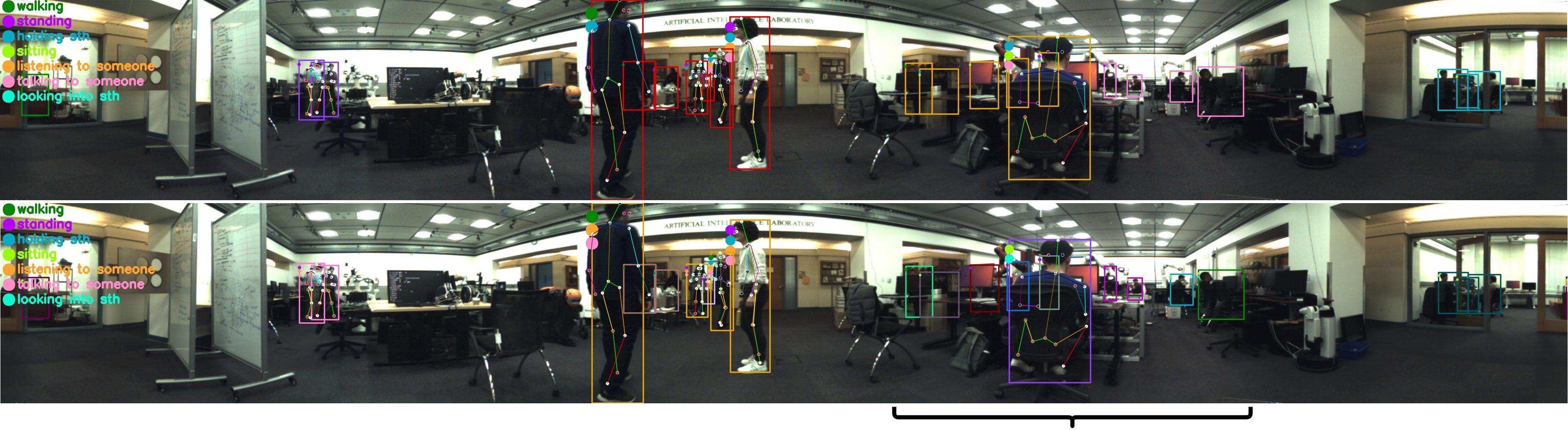}
   \vspace{-1.2em}
   \caption{A failure case of S-MAE prediction in the social grouping task. Since S-MAE relies on multi-person joints trajectory to predict the social groups, it predicts people that have similar joints movement and are close to each other in 2D as one social group.}
   \label{fig:v2}
   \end{figure*}

Further, as shown in Fig.~\ref{fig:v2}, although the predicted social groups and action labels are correct compared to the ground-truth in most of the cases, a failure case of the S-MAE model is indicated by the arrow. S-MAE has predicted all the individuals in that part of the sub-figure as one social group since their joints' trajectory is still and are close to each other in 2D. We believe that a future direction to improve social grouping is to incorporate 3D data.

\begin{figure*}[t]
\vspace{-.5em}
   \centering
   \includegraphics[width=\textwidth]{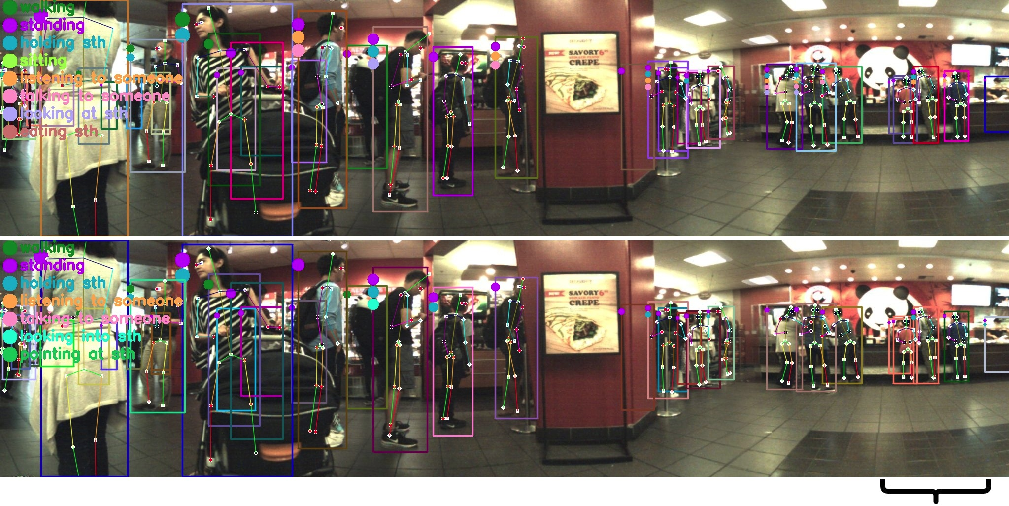}
   \hfill
   \includegraphics[width=\textwidth]{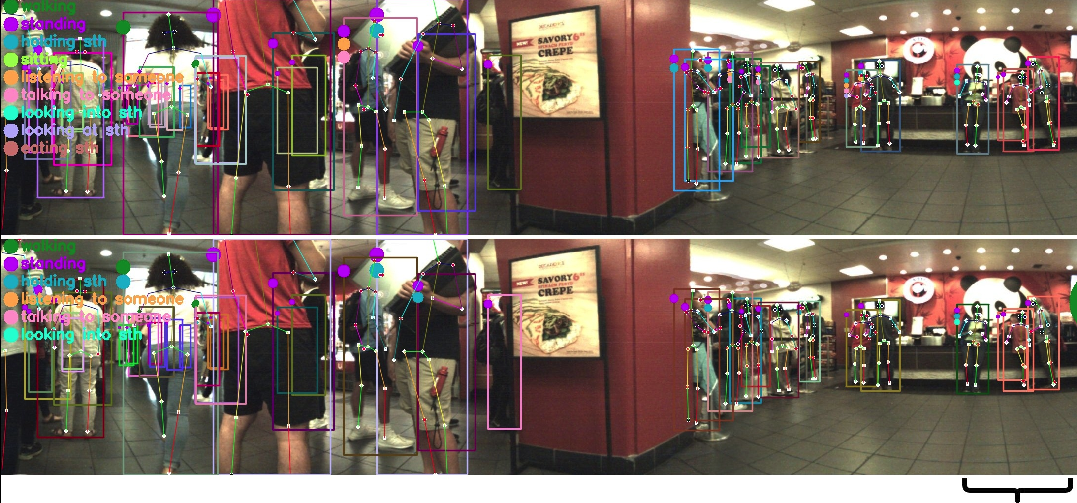}
   \vspace{-.9em}
   \caption{2 different samples of S-MAE predictions on crowded and complex scenarios for social grouping and action understanding tasks. Although it is hard to imply whether all customers are interacting with each other and with the waiter, in the first sample, the model has predicted the closer customer to the waiter and the waiter as one social group. In the second sample, the model has predicted both customers and the waiter as one social group, highlighted by the arrow.}
   \vspace{-.9em}
   \label{fig:v3}
   \end{figure*}

JRDB-Act~\cite{ehsanpour2022jrdb} contains crowded footage and complex interaction scenarios as shown in Fig.~\ref{fig:v3}. S-MAE performs qualitatively well on such sequences. For instance, in the case highlighted by the arrow in which two customers are interacting with the waiter, the model has predicted the customer closer to the waiter as one social group and the other customer as another social group. Although it is hard to recognize whether both customers are interacting with the waiter, in the second case, the model has correctly predicted both customers and the waiter as one social group.

%% file: main.bbl
\begin{thebibliography}{10}
\providecommand{\url}[1]{\texttt{#1}}
\providecommand{\urlprefix}{URL }
\providecommand{\doi}[1]{https://doi.org/#1}

\bibitem{adeli2020socially}
Adeli, V., Adeli, E., Reid, I., Niebles, J.C., Rezatofighi, H.: Socially and contextually aware human motion and pose forecasting. IEEE Robotics and Automation Letters  \textbf{5}(4),  6033--6040 (2020)

\bibitem{adeli2021tripod}
Adeli, V., Ehsanpour, M., Reid, I., Niebles, J.C., Savarese, S., Adeli, E., Rezatofighi, H.: Tripod: Human trajectory and pose dynamics forecasting in the wild. In: Proceedings of the IEEE/CVF International Conference on Computer Vision. pp. 13390--13400 (2021)

\bibitem{assran2021semi}
Assran, M., Caron, M., Misra, I., Bojanowski, P., Joulin, A., Ballas, N., Rabbat, M.: Semi-supervised learning of visual features by non-parametrically predicting view assignments with support samples. In: Proceedings of the IEEE/CVF International Conference on Computer Vision. pp. 8443--8452 (2021)

\bibitem{bao2021beit}
Bao, H., Dong, L., Piao, S., Wei, F.: Beit: Bert pre-training of image transformers. arXiv preprint arXiv:2106.08254  (2021)

\bibitem{baradel2022posebert}
Baradel, F., Br{\'e}gier, R., Groueix, T., Weinzaepfel, P., Kalantidis, Y., Rogez, G.: Posebert: A generic transformer module for temporal 3d human modeling. IEEE Transactions on Pattern Analysis and Machine Intelligence  (2022)

\bibitem{brown2020language}
Brown, T., Mann, B., Ryder, N., Subbiah, M., Kaplan, J.D., Dhariwal, P., Neelakantan, A., Shyam, P., Sastry, G., Askell, A., et~al.: Language models are few-shot learners. Advances In Neural Information Processing Systems  \textbf{33},  1877--1901 (2020)

\bibitem{cao2017realtime}
Cao, Z., Simon, T., Wei, S.E., Sheikh, Y.: Realtime multi-person 2d pose estimation using part affinity fields. In: Proceedings of the IEEE/CVF International Conference on Computer Vision. pp. 7291--7299 (2017)

\bibitem{caron2021emerging}
Caron, M., Touvron, H., Misra, I., J{\'e}gou, H., Mairal, J., Bojanowski, P., Joulin, A.: Emerging properties in self-supervised vision transformers. In: Proceedings of the IEEE/CVF International Conference on Computer Vision. pp. 9650--9660 (2021)

\bibitem{carreira2017quo}
Carreira, J., Zisserman, A.: Quo vadis, action recognition? a new model and the kinetics dataset. In: Proceedings of the IEEE/CVF International Conference on Computer Vision. pp. 6299--6308 (2017)

\bibitem{chen2022graph}
Chen, H., Zhang, S., Xu, G.: Graph masked autoencoder. arXiv preprint arXiv:2202.08391  (2022)

\bibitem{Chen_2020_CVPR}
Chen, L., Ai, H., Chen, R., Zhuang, Z., Liu, S.: Cross-view tracking for multi-human 3d pose estimation at over 100 fps. In: The IEEE/CVF Conference on Computer Vision and Pattern Recognition (2020)

\bibitem{chen2020generative}
Chen, M., Radford, A., Child, R., Wu, J., Jun, H., Luan, D., Sutskever, I.: Generative pretraining from pixels. In: International Conference on Machine Learning. pp. 1691--1703 (2020)

\bibitem{chen2020simple}
Chen, T., Kornblith, S., Norouzi, M., Hinton, G.: A simple framework for contrastive learning of visual representations. In: International Conference on Machine Learning. pp. 1597--1607 (2020)

\bibitem{choi2014discovering}
Choi, W., Chao, Y.W., Pantofaru, C., Savarese, S.: Discovering groups of people in images. In: Proceedings of The European Conference on Computer Vision. pp. 417--433 (2014)

\bibitem{deng2009imagenet}
Deng, J., Dong, W., Socher, R., Li, L.J., Li, K., Fei-Fei, L.: Imagenet: A large-scale hierarchical image database. In: Proceedings of the IEEE/CVF International Conference on Computer Vision. pp. 248--255 (2009)

\bibitem{devlin2018bert}
Devlin, J., Chang, M.W., Lee, K., Toutanova, K.: Bert: Pre-training of deep bidirectional transformers for language understanding. arXiv preprint arXiv:1810.04805  (2018)

\bibitem{dong2021peco}
Dong, X., Bao, J., Zhang, T., Chen, D., Zhang, W., Yuan, L., Chen, D., Wen, F., Yu, N.: Peco: Perceptual codebook for bert pre-training of vision transformers. arXiv preprint arXiv:2111.12710  (2021)

\bibitem{dosovitskiy2020image}
Dosovitskiy, A., Beyer, L., Kolesnikov, A., Weissenborn, D., Zhai, X., Unterthiner, T., Dehghani, M., Minderer, M., Heigold, G., Gelly, S., et~al.: An image is worth 16x16 words: Transformers for image recognition at scale. arXiv preprint arXiv:2010.11929  (2020)

\bibitem{ehsanpour2020joint}
Ehsanpour, M., Abedin, A., Saleh, F., Shi, J., Reid, I., Rezatofighi, H.: Joint learning of social groups, individuals action and sub-group activities in videos. In: Proceedings of The European Conference on Computer Vision. pp. 177--195 (2020)

\bibitem{ehsanpour2022jrdb}
Ehsanpour, M., Saleh, F., Savarese, S., Reid, I., Rezatofighi, H.: Jrdb-act: A large-scale dataset for spatio-temporal action, social group and activity detection. In: Proceedings of the IEEE/CVF Conference on Computer Vision and Pattern Recognition. pp. 20983--20992 (2022)

\bibitem{fang2017rmpe}
Fang, H.S., Xie, S., Tai, Y.W., Lu, C.: Rmpe: Regional multi-person pose estimation. In: Proceedings of the IEEE/CVF International Conference on Computer Vision. pp. 2334--2343 (2017)

\bibitem{feichtenhofer2022masked}
Feichtenhofer, C., Li, Y., He, K., et~al.: Masked autoencoders as spatiotemporal learners. Advances in Neural Information Processing Systems  \textbf{35},  35946--35958 (2022)

\bibitem{fu2023improved}
Fu, J., Dang, Y., Yin, R., Zhang, S., Zhou, F., Zhao, W., Yin, J.: An improved baseline framework for pose estimation challenge at eccv 2022 visual perception for navigation in human environments workshop. arXiv preprint arXiv:2303.07141  (2023)

\bibitem{ge2012vision}
Ge, W., Collins, R.T., Ruback, R.B.: Vision-based analysis of small groups in pedestrian crowds. IEEE Transactions on Pattern Analysis and Machine Intelligence  \textbf{34}(5),  1003--1016 (2012)

\bibitem{gu2018ava}
Gu, C., Sun, C., Ross, D.A., Vondrick, C., Pantofaru, C., Li, Y., Vijayanarasimhan, S., Toderici, G., Ricco, S., Sukthankar, R., et~al.: Ava: A video dataset of spatio-temporally localized atomic visual actions. In: Proceedings of the IEEE/CVF International Conference on Computer Vision. pp. 6047--6056 (2018)

\bibitem{guler2018densepose}
G{\"u}ler, R.A., Neverova, N., Kokkinos, I.: Densepose: Dense human pose estimation in the wild. In: Proceedings of the IEEE/CVF International Conference on Computer Vision. pp. 7297--7306 (2018)

\bibitem{guo2022multi}
Guo, W., Bie, X., Alameda-Pineda, X., Moreno-Noguer, F.: Multi-person extreme motion prediction. In: Proceedings of the IEEE/CVF Conference on Computer Vision and Pattern Recognition. pp. 13053--13064 (2022)

\bibitem{gupta2018social}
Gupta, A., Johnson, J., Fei-Fei, L., Savarese, S., Alahi, A.: Social gan: Socially acceptable trajectories with generative adversarial networks. In: Proceedings of the IEEE/CVF International Conference on Computer Vision. pp. 2255--2264 (2018)

\bibitem{han2022panoramic}
Han, R., Yan, H., Li, J., Wang, S., Feng, W., Wang, S.: Panoramic human activity recognition. In: Proceedings of The European Conference on Computer Vision. pp. 244--261 (2022)

\bibitem{he2022masked}
He, K., Chen, X., Xie, S., Li, Y., Doll{\'a}r, P., Girshick, R.: Masked autoencoders are scalable vision learners. In: Proceedings of the IEEE/CVF Conference on Computer Vision and Pattern Recognition. pp. 16000--16009 (2022)

\bibitem{he2021know}
He, Y., Yu, W., Han, J., Wei, X., Hong, X., Gong, Y.: Know your surroundings: Panoramic multi-object tracking by multimodality collaboration. In: Proceedings of the IEEE/CVF Conference on Computer Vision and Pattern Recognition. pp. 2969--2980 (2021)

\bibitem{hou2022graphmae}
Hou, Z., Liu, X., Cen, Y., Dong, Y., Yang, H., Wang, C., Tang, J.: Graphmae: Self-supervised masked graph autoencoders. In: Proceedings of the 28th ACM SIGKDD Conference on Knowledge Discovery and Data Mining. pp. 594--604 (2022)

\bibitem{hung2011detecting}
Hung, H., Kr{\"o}se, B.: Detecting f-formations as dominant sets. In: Proceedings of The International Conference on Multimodal Interfaces. pp. 231--238 (2011)

\bibitem{jahangard2023real}
Jahangard, S., Hayat, M., Rezatofighi, H.: Real-time trajectory-based social group detection. arXiv preprint arXiv:2304.05678  (2023)

\bibitem{jiang2022dual}
Jiang, J., Chen, J., Guo, Y.: A dual-masked auto-encoder for robust motion capture with spatial-temporal skeletal token completion. In: Proceedings of the 30th ACM International Conference on Multimedia. pp. 5123--5131 (2022)

\bibitem{kosaraju2019social}
Kosaraju, V., Sadeghian, A., Mart{\'\i}n-Mart{\'\i}n, R., Reid, I., Rezatofighi, H., Savarese, S.: Social-bigat: Multimodal trajectory forecasting using bicycle-gan and graph attention networks. Advances in Neural Information Processing Systems  \textbf{32} (2019)

\bibitem{kothari2021interpretable}
Kothari, P., Sifringer, B., Alahi, A.: Interpretable social anchors for human trajectory forecasting in crowds. In: Proceedings of the IEEE/CVF Conference on Computer Vision and Pattern Recognition. pp. 15556--15566 (2021)

\bibitem{li2022self}
Li, J., Han, R., Yan, H., Qian, Z., Feng, W., Wang, S.: Self-supervised social relation representation for human group detection. In: Proceedings of The European Conference on Computer Vision. pp. 142--159 (2022)

\bibitem{li2022maskgae}
Li, J., Wu, R., Sun, W., Chen, L., Tian, S., Zhu, L., Meng, C., Zheng, Z., Wang, W.: Maskgae: masked graph modeling meets graph autoencoders. arXiv preprint arXiv:2205.10053  (2022)

\bibitem{liang2022meshmae}
Liang, Y., Zhao, S., Yu, B., Zhang, J., He, F.: Meshmae: Masked autoencoders for 3d mesh data analysis. In: Proceedings of The European Conference on Computer Vision. pp. 37--54 (2022)

\bibitem{liu2022masked}
Liu, H., Cai, M., Lee, Y.J.: Masked discrimination for self-supervised learning on point clouds. In: Proceedings of The European Conference on Computer Vision. pp. 657--675 (2022)

\bibitem{luo2020arbee}
Luo, Y., Ye, J., Adams, R.B., Li, J., Newman, M.G., Wang, J.Z.: Arbee: Towards automated recognition of bodily expression of emotion in the wild. International Journal of Computer Vision  \textbf{128},  1--25 (2020)

\bibitem{mahmood2019amass}
Mahmood, N., Ghorbani, N., Troje, N.F., Pons-Moll, G., Black, M.J.: Amass: Archive of motion capture as surface shapes. In: Proceedings of the IEEE/CVF International Conference on Computer Vision. pp. 5442--5451 (2019)

\bibitem{mangalam2020not}
Mangalam, K., Girase, H., Agarwal, S., Lee, K.H., Adeli, E., Malik, J., Gaidon, A.: It is not the journey but the destination: Endpoint conditioned trajectory prediction. In: Proceedings of The European Conference on Computer Vision. pp. 759--776 (2020)

\bibitem{mao2020history}
Mao, W., Liu, M., Salzmann, M.: History repeats itself: Human motion prediction via motion attention. In: Proceedings of The European Conference on Computer Vision. pp. 474--489 (2020)

\bibitem{mao2019learning}
Mao, W., Liu, M., Salzmann, M., Li, H.: Learning trajectory dependencies for human motion prediction. In: Proceedings of the IEEE/CVF International Conference on Computer Vision. pp. 9489--9497 (2019)

\bibitem{martin2021jrdb}
Martin-Martin, R., Patel, M., Rezatofighi, H., Shenoi, A., Gwak, J., Frankel, E., Sadeghian, A., Savarese, S.: Jrdb: A dataset and benchmark of egocentric robot visual perception of humans in built environments. IEEE Transactions on Pattern Analysis and Machine Intelligence  (2021)

\bibitem{mehta2018single}
Mehta, D., Sotnychenko, O., Mueller, F., Xu, W., Sridhar, S., Pons-Moll, G., Theobalt, C.: Single-shot multi-person 3d pose estimation from monocular rgb. In: International Conference on 3D Vision. pp. 120--130 (2018)

\bibitem{min2022voxel}
Min, C., Zhao, D., Xiao, L., Nie, Y., Dai, B.: Voxel-mae: Masked autoencoders for pre-training large-scale point clouds. arXiv preprint arXiv:2206.09900  (2022)

\bibitem{pang2022masked}
Pang, Y., Wang, W., Tay, F.E., Liu, W., Tian, Y., Yuan, L.: Masked autoencoders for point cloud self-supervised learning. In: Proceedings of The European Conference on Computer Vision. pp. 604--621. Springer (2022)

\bibitem{patron2010high}
Patron-Perez, A., Marszalek, M., Zisserman, A., Reid, I.: High five: Recognising human interactions in tv shows. In: British Machine Vision Conference. vol.~1, p.~33 (2010)

\bibitem{pishchulin2016deepcut}
Pishchulin, L., Insafutdinov, E., Tang, S., Andres, B., Andriluka, M., Gehler, P.V., Schiele, B.: Deepcut: Joint subset partition and labeling for multi person pose estimation. In: Proceedings of the IEEE/CVF International Conference on Computer Vision. pp. 4929--4937 (2016)

\bibitem{radford2019language}
Radford, A., Wu, J., Child, R., Luan, D., Amodei, D., Sutskever, I., et~al.: Language models are unsupervised multitask learners. OpenAI blog  \textbf{1}(8), ~9 (2019)

\bibitem{ramesh2021zero}
Ramesh, A., Pavlov, M., Goh, G., Gray, S., Voss, C., Radford, A., Chen, M., Sutskever, I.: Zero-shot text-to-image generation. In: International Conference on Machine Learning. pp. 8821--8831 (2021)

\bibitem{sadeghian2019sophie}
Sadeghian, A., Kosaraju, V., Sadeghian, A., Hirose, N., Rezatofighi, H., Savarese, S.: Sophie: An attentive gan for predicting paths compliant to social and physical constraints. In: Proceedings of the IEEE/CVF International Conference on Computer Vision. pp. 1349--1358 (2019)

\bibitem{setti2013multi}
Setti, F., Lanz, O., Ferrario, R., Murino, V., Cristani, M.: Multi-scale f-formation discovery for group detection. In: IEEE International Conference on Image Processing. pp. 3547--3551 (2013)

\bibitem{sun2020recursive}
Sun, J., Jiang, Q., Lu, C.: Recursive social behavior graph for trajectory prediction. In: Proceedings of the IEEE/CVF International Conference on Computer Vision. pp. 660--669 (2020)

\bibitem{swofford2020improving}
Swofford, M., Peruzzi, J., Tsoi, N., Thompson, S., Mart{\'\i}n-Mart{\'\i}n, R., Savarese, S., V{\'a}zquez, M.: Improving social awareness through dante: Deep affinity network for clustering conversational interactants. Proceedings of the ACM on Human-Computer Interaction  \textbf{4},  1--23 (2020)

\bibitem{tan2021vimpac}
Tan, H., Lei, J., Wolf, T., Bansal, M.: Vimpac: Video pre-training via masked token prediction and contrastive learning. arXiv preprint arXiv:2106.11250  (2021)

\bibitem{tan2022mgae}
Tan, Q., Liu, N., Huang, X., Chen, R., Choi, S.H., Hu, X.: Mgae: Masked autoencoders for self-supervised learning on graphs. arXiv preprint arXiv:2201.02534  (2022)

\bibitem{tong2022videomae}
Tong, Z., Song, Y., Wang, J., Wang, L.: Videomae: Masked autoencoders are data-efficient learners for self-supervised video pre-training. arXiv preprint arXiv:2203.12602  (2022)

\bibitem{tsao2022social}
Tsao, L.W., Wang, Y.K., Lin, H.S., Shuai, H.H., Wong, L.K., Cheng, W.H.: Social-ssl: Self-supervised cross-sequence representation learning based on transformers for multi-agent trajectory prediction. In: Proceedings of The European Conference on Computer Vision. pp. 234--250 (2022)

\bibitem{vendrow2022somoformer}
Vendrow, E., Kumar, S., Adeli, E., Rezatofighi, H.: Somoformer: Multi-person pose forecasting with transformers. arXiv preprint arXiv:2208.14023  (2022)

\bibitem{vincent2008extracting}
Vincent, P., Larochelle, H., Bengio, Y., Manzagol, P.A.: Extracting and composing robust features with denoising autoencoders. In: Proceedings of the International Conference on Machine Learning. pp. 1096--1103 (2008)

\bibitem{vincent2010stacked}
Vincent, P., Larochelle, H., Lajoie, I., Bengio, Y., Manzagol, P.A., Bottou, L.: Stacked denoising autoencoders: Learning useful representations in a deep network with a local denoising criterion. Journal of machine learning research  \textbf{11}(12) (2010)

\bibitem{von2018recovering}
Von~Marcard, T., Henschel, R., Black, M.J., Rosenhahn, B., Pons-Moll, G.: Recovering accurate 3d human pose in the wild using imus and a moving camera. In: Proceedings of The European Conference on Computer Vision. pp. 601--617 (2018)

\bibitem{wang2021simple}
Wang, C., Wang, Y., Huang, Z., Chen, Z.: Simple baseline for single human motion forecasting. In: Proceedings of the IEEE/CVF International Conference on Computer Vision. pp. 2260--2265 (2021)

\bibitem{wang2021multi}
Wang, J., Xu, H., Narasimhan, M., Wang, X.: Multi-person 3d motion prediction with multi-range transformers. Advances in Neural Information Processing Systems  \textbf{34},  6036--6049 (2021)

\bibitem{wang2020deep}
Wang, J., Sun, K., Cheng, T., Jiang, B., Deng, C., Zhao, Y., Liu, D., Mu, Y., Tan, M., Wang, X., et~al.: Deep high-resolution representation learning for visual recognition. IEEE Transactions on Pattern Analysis and Machine Intelligence  \textbf{43}(10),  3349--3364 (2020)

\bibitem{wang2022bevt}
Wang, R., Chen, D., Wu, Z., Chen, Y., Dai, X., Liu, M., Jiang, Y.G., Zhou, L., Yuan, L.: Bevt: Bert pretraining of video transformers. In: Proceedings of the IEEE/CVF Conference on Computer Vision and Pattern Recognition. pp. 14733--14743 (2022)

\bibitem{wei2022masked}
Wei, C., Fan, H., Xie, S., Wu, C.Y., Yuille, A., Feichtenhofer, C.: Masked feature prediction for self-supervised visual pre-training. In: Proceedings of the IEEE/CVF Conference on Computer Vision and Pattern Recognition. pp. 14668--14678 (2022)

\bibitem{xie2022simmim}
Xie, Z., Zhang, Z., Cao, Y., Lin, Y., Bao, J., Yao, Z., Dai, Q., Hu, H.: Simmim: A simple framework for masked image modeling. In: Proceedings of the IEEE/CVF Conference on Computer Vision and Pattern Recognition. pp. 9653--9663 (2022)

\bibitem{xustochastic}
Xu, S., Wang, Y.X., Gui, L.: Stochastic multi-person 3d motion forecasting. In: The Eleventh International Conference on Learning Representations (2023)

\bibitem{xu2022vitpose}
Xu, Y., Zhang, J., Zhang, Q., Tao, D.: Vitpose: Simple vision transformer baselines for human pose estimation. arXiv preprint arXiv:2204.12484  (2022)

\bibitem{zhang2022point}
Zhang, R., Guo, Z., Gao, P., Fang, R., Zhao, B., Wang, D., Qiao, Y., Li, H.: Point-m2ae: multi-scale masked autoencoders for hierarchical point cloud pre-training. arXiv preprint arXiv:2205.14401  (2022)

\bibitem{zhou2021ibot}
Zhou, J., Wei, C., Wang, H., Shen, W., Xie, C., Yuille, A., Kong, T.: ibot: Image bert pre-training with online tokenizer. arXiv preprint arXiv:2111.07832  (2021)

\end{thebibliography}
